\documentclass[sigconf]{acmart}
\usepackage{subfigure}
\usepackage{bm}
\usepackage{multirow}
\usepackage{balance}

\copyrightyear{2020}
\acmYear{2020}
\setcopyright{iw3c2w3}
\acmConference[WWW '20]{Proceedings of The Web Conference 2020}{April 20--24, 2020}{Taipei, Taiwan}
\acmBooktitle{Proceedings of The Web Conference 2020 (WWW '20), April 20--24, 2020, Taipei, Taiwan}
\acmPrice{}
\acmDOI{10.1145/3366423.3380172}
\acmISBN{978-1-4503-7023-3/20/04}
\begin{document}

%\title{Modeling Users' Behavior Sequences with Hierarchical Explainable Network for Cross-domain Fraud Detection}
\title[Hierarchical Explainable Network for Cross-domain Fraud Detection]{Modeling Users' Behavior Sequences with Hierarchical Explainable Network for Cross-domain Fraud Detection}

\author{Yongchun Zhu$^{1,2,3}$, Dongbo Xi$^{1,2,3}$, Bowen Song$^{3}$, Fuzhen Zhuang$^{1,2}$,}
\author{Shuai Chen$^{3}$, Xi Gu$^3$ and Qing He$^{1,2}$}
\affiliation{%
 \institution{$^1$Key Lab of Intelligent Information Processing of Chinese Academy of Sciences (CAS), Institute of Computing Technology, CAS, Beijing 100190, China}} 
\affiliation{%
\institution{$^2$University of Chinese Academy of Sciences, Beijing 100049, China}} 
\affiliation{%
\institution{$^3$Alipay (Hangzhou) Information \& Technology Co., Ltd.}} 
\affiliation{\{zhuyongchun18s, xidongbo17s, zhuangfuzhen, heqing\}@ict.ac.cn, \{bowen.sbw, shuai.cs, guxi.gx\}@antfin.com}
\thanks{*Fuzhen Zhuang and Bowen Song are corresponding authors.}

\renewcommand{\shortauthors}{Y. Zhu et al.}

\begin{abstract}
    With the explosive growth of the e-commerce industry, detecting online transaction fraud in real-world applications has become increasingly important to the development of e-commerce platforms. The sequential behavior history of users provides useful information in differentiating fraudulent payments from regular ones. Recently, some approaches have been proposed to solve this sequence-based fraud detection problem. However, these methods usually suffer from two problems: the prediction results are difficult to explain and the exploitation of the internal information of behaviors is insufficient. To tackle the above two problems,  we propose a \textit{Hierarchical Explainable Network} (HEN) to model users' behavior sequences, which could not only improve the performance of fraud detection but also make the inference process interpretable.

    Meanwhile, as e-commerce business expands to new domains, e.g., new countries or new markets, one major problem for modeling user behavior in fraud detection systems is the limitation of data collection, e.g., very few data/labels available. Thus, in this paper, we further propose a transfer framework to tackle the cross-domain fraud detection problem, which aims to transfer knowledge from existing domains (source domains) with enough and mature data to improve the performance in the new domain (target domain). Our proposed method is a general transfer framework that could not only be applied upon HEN but also various existing models in the Embedding \& MLP paradigm.
    
    By utilizing data from a world-leading cross-border e-commerce platform, we conduct extensive experiments in detecting card-stolen transaction frauds in different countries to demonstrate the superior performance of HEN. Besides, based on 90 transfer task experiments, we also demonstrate that our transfer framework could not only contribute to the cross-domain fraud detection task with HEN, but also be universal and expandable for various existing models. Moreover, HEN and the transfer framework form three-level attention which greatly increases the explainability of the detection results.
\end{abstract}





\maketitle

\section{Introduction}\label{sec:intro}
With the rapid growth of information technologies, e-commerce has become prevalent nowadays. A large online e-commerce website serves millions of users with numerous products and services everyday, providing them a convenient, fast, and reliable manner of shopping, service acquisition, reviewing, comment feedback, etc. Unfortunately, the problem of online transaction fraud has become increasingly prominent, putting the finance of e-commerce at risk~\cite{fastoso2012risk}. Online fraud activities have caused a loss in billions of dollars~\footnote{https://www.signifyd.com/blog/2017/10/26/ecommerce-fraud-eight-industries/}.

\begin{figure}[h]
	\centering
	\begin{minipage}[b]{1\linewidth}
		\centering
		\includegraphics[width=.95\linewidth]{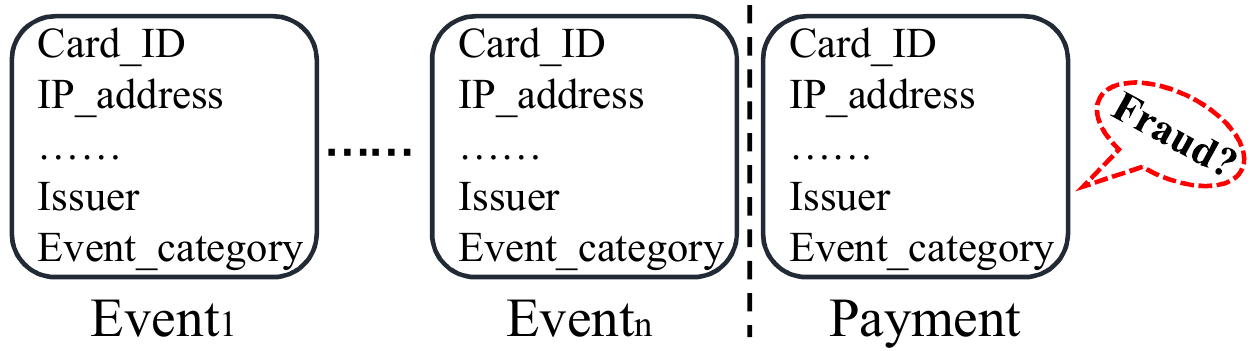}
	\end{minipage}
	\caption{The sequence-based fraud detection task which exploits the historical behavior sequence to help the prediction of the target payment event. We have 6 classes of events (e.g., sign up, sign in, payment, etc) and the target event is a payment event.}\label{fig:sequence}
\end{figure}

Detection of real-time online fraud is critical to the development of e-commerce platforms. To solve the fraud detection problem, many approaches have been proposed~\cite{phua2010comprehensive,cohen1995fast,baulier2000automated,brause1999neural,fu2016credit,zakaryazad2016profit}. However, these approaches did not consider the sequential behavior history of users, which has been utilized in many studies~\cite{wang2017session,jurgovsky2018sequence} to improve the performance of the fraud detection system. Such sequence prediction task exploits the users' historical behavior sequences to help differentiate fraudulent payments from regular ones, as shown in Figure~\ref{fig:sequence}. For the sequence prediction task, some effective models have been proposed to capture the sequential information in user's behavior history, such as Markov Chains based methods~\cite{zhang2018sequential}, convolutional neural networks based methods~\cite{tang2018personalized}, and recurrent neural networks based methods~\cite{wang2017session,jurgovsky2018sequence}. These methods utilize sequential information to further improve performance. However, (1) The prediction results are difficult to explain for these methods. (2) They focus more on the sequential information of the behaviors, but fail to thoroughly exploit the internal information of each behavior, e.g., only the first-order information of fields' embeddings is used to represent events~\cite{wang2017session,zhang2018sequential,jurgovsky2018sequence}.

In order to tackle the above two problems, we propose a \textit{Hierarchical Explainable Network} (HEN) to model users' behavior sequences, which could not only improve the performance of fraud detection but also help answer this ``why''. HEN contains two-level extractors: 1) Field-level extractor extracts the representations of behavior events that contain both first- and second-order information from the embedding of fields, and it learns to choose informative fields, e.g., the card-related fields would be more important than the time-related fields. 2) Event-level extractor extracts the representations of users' historical behavior sequences from the representations of behavior events and could score the event importance. Besides, the wide layer of HEN could help to identify the specific value of fields with high-risk/low-risk, which could be used as whitelist/blacklist.

In this paper, the fraud detection dataset is collected from one of the world-leading cross-border e-commerce companies. As its e-commerce business expands to new domains, e.g., new countries or new markets, one major problem for modeling users' behavior in fraud detection systems is the limitation of data collection, e.g., very few data/labels available.

Hence, we introduce cross-domain fraud detection, which aims to transfer knowledge from existing domains (source domains) with enough and mature data to improve the performance in the new domain (target domain). The main challenges of the cross-domain fraud detection problem are elaborated as follows: (1) In our problem, the source and target domains share some knowledge but also have some specific characteristics, e.g., the IP address of different countries are different, while the card type is shared.  (2) For the prediction of different samples, the weight of the shared and specific knowledge would be different. (3) Due to the domain-specific knowledge, the distributions of source and target samples can differ in many ways.

Due to these challenges, modeling users' behavior sequences with a single structure, as shown in most previous cross-domain works~\cite{long2015learning,ganin2016domain,sun2016deep,xie2018learning}, could not meet the demand of our fraud detection task, because it only learns shared representations without paying attention to domain-specific knowledge. 
To tackle the cross-domain fraud detection problem, we propose three important aspects: (1) The network should be divided into two parts to capture domain-shared and domain-specific knowledge, respectively. (2) Domain attention is useful to automatically learn the weights of domain-shared and domain-specific representations for each sample. (3) Existing marginal~\cite{long2015learning,ganin2016domain,sun2016deep} and conditional~\cite{long2013transfer,xie2018learning} alignment methods are not suitable for the cross-domain fraud detection tasks which suffer extremely unbalanced categories problem. To this end, we propose Class-aware Euclidean Distance which considers both intra- and inter-class information. Along this line, we propose a general transfer framework that can be applied upon various existing models in the Embedding \& MLP paradigm.

The main contributions of this work are summarized into three folds:
\begin{itemize}
    \item To solve the sequence-based fraud detection problem, we propose hierarchical explainable network (HEN) to model users' behavior sequences, which could not only improve the performance of fraud detection but also give reasonable explanations for the prediction results.
    \item For cross-domain fraud detection, we propose a general transfer framework that can be applied upon various existing models in the Embedding \& MLP paradigm. 
    \item We perform experiments on real-world datasets of four countries to demonstrate the effectiveness of HEN. In addition, we demonstrate our transfer framework is general for various existing models on 90 transfer tasks. Finally, we conduct case study to prove the explainability of HEN and the transfer framework.
\end{itemize}

\section{Related Work}\label{sec:relatedWork}
In this section, we will introduce the related work from three aspects: Fraud Detection, Sequence Prediction, Transfer Learning.

Fraud Detection:  Researchers have investigated in fraud detection problems for a long time. Early explorations of fraud detection focus on rule-based methods. Quinlan~\cite{quinlan1990learning} and Cohen~\cite{cohen1995fast} introduced assertion statement of IF \{conditions\} and THEN \{a consequent\} to recognize fraud records. Association rules have been applied to detect credit card fraud~\cite{brause1999neural}. Rosset et al.~\cite{rosset1999discovery} presented a two-stage rules-based fraud detection system to detect telephone fraud. 

However, fraudulent behaviors change over time, which greatly deteriorates the effectiveness of rules summarized by expert experience. Recently, with millions of transaction data available, more and more data-driven and learning-based methods are applied for the fraud detection problem. SVM-based ensemble strategy was utilized for detecting telecommunication subscription fraud and credit fraud~\cite{wang2012hybrid}. Some fraud detection works focused on using graphs for spotting frauds~\cite{tian2015crowd,tseng2015fraudetector}. Convolutional neural network (CNN) has been applied for credit card fraud detection~\cite{fu2016credit}. Some works used recurrent neural networks for sequence-based fraud detection~\cite{wang2017session,zhang2018sequential,jurgovsky2018sequence}. In this paper, we also focus on sequence-based fraud detection, and compare our HEN with the RNN based methods in experiments.

Sequence Prediction: Sequence prediction~\cite{li2018learning,huang2019ekt} is a kind of prediction problem which exploits the users' historical behavior sequences to help prediction. He et al.~\cite{he2016fusing} combined similarity-based models with high-order Markov chains to make personalized sequential recommendations. Tang et al.~\cite{tang2018personalized} proposed to apply the convolutional neural network (CNN) on the embedding sequence, where the short-term contexts can be captured by the convolutional operations.  Some works~\cite{wang2017session,zhang2018sequential,jurgovsky2018sequence} exploited recurrent neural networks for sequence-based fraud detection. Zhou et al.~\cite{zhou2018deep} exploited an attention module for sequence recommendation.
Besides, a method called Multi-temporal-range Mixture Model (M3)~\cite{tang2019towards} has been proposed to apply a dense layer to extract behavior representation from the embedding of fields, and then employ a mixture of models to deal with both short-term and long-term dependencies. Nevertheless, these studies focus on the sequential information without effectively exploiting the internal information of each behavior. In addition, most of these methods ignore the importance of explainability while our HEN is able to give reasonable explanations for the prediction results.

Transfer Learning: Transfer learning aims to leverage knowledge from a source domain to improve the learning performance or minimize the number of labeled examples required in a target domain~\cite{pan2009survey,zhuang2019comprehensive}. Some transfer methods have been widely adopted for many problems, such as pre-training a model on a large dataset and then fine-tuning on the target task~\cite{hinton2006fast}. However, the improvement of such general transfer methods is limited. Recently, transfer learning on computer vision and natural language process has attracted the attention of amounts of researchers~\cite{long2015learning,zhuang2015supervised,ganin2016domain,sun2016deep,wang2017balanced}. These approaches largely improve performance, but most of them have not been used for practical industrial problems. There are some transfer approaches for practical industrial problems, such as cross-domain recommendation~\cite{hu2019transfer}, adaptively handwritten Chinese character recognition~\cite{zhu2019adaptively}, passenger demand forecasting~\cite{bai2019passenger}. However, to the best of our knowledge, this is the first work to propose cross-domain fraud detection. 
\begin{figure*}[h]
	\centering
	\begin{minipage}[b]{1\linewidth}
		\centering
		\includegraphics[width=.80\linewidth]{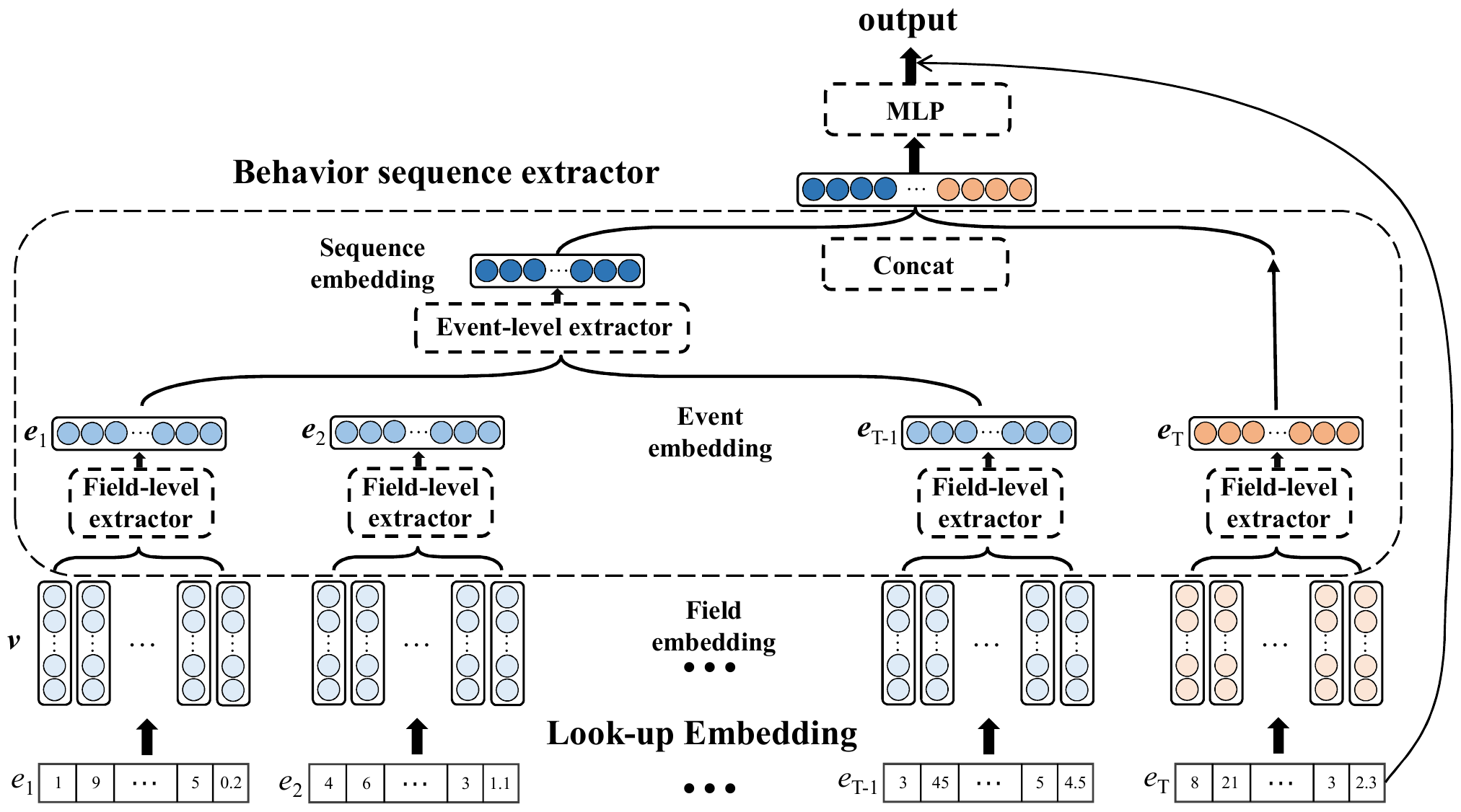}
	\end{minipage}
	\caption{The structure of hierarchical explainable network.}\label{fig:HEN}
\end{figure*}
\section{Hierarchical Explainable Network}
\subsection{Problem Statement}\label{problemdef}
Given a user's behavior event sequence $E = [e_1, e_2, \cdots, e_T]$, where $T$ is the the length of the sequence. Each behavior event $e$ has $n$ fields, such as IP\_address, Event\_category, Issuer, etc. The behavior event is denoted as $e_t = [x^t_1, x^t_2,\cdots,x_n^t], (1\leq t \leq T)$, where $x^t_i$ denotes the value of the $i$-th field. The task is to predict whether the target payment event $e_T$ is fraud ($y = 1$ denotes $e_T$ is fraud) with the user's historical behavior event sequence $[e_1, e_2, \cdots, e_{T-1}]$ and available information of the target event $e_T$. In such setting, the task can be formulated as a binary prediction task. 

\subsection{Look-up Embedding}\label{lookup}
Look-up embedding has been widely adopted to learn dense representations from raw data for prediction~\cite{he2017neural,tang2018personalized}. In practice, we have two types of fields, categorical fields that have a limited number of distinct values (such as Issuer, Event\_category) and numerical fields which are continuous values (such as Balance\_amount). For the two types of fields, the methods of look-up embedding are different. We formulate the embedding matrix or the look-up table of the $i$-th field as:
\begin{equation}
    \Phi_i \in
\begin{cases}
\mathbb{R}^{m \times k}& \text{the } i\text{-th field is categorical field}\\
\mathbb{R}^{k}& \text{the } i\text{-th field is numerical field}
\end{cases},
\end{equation}
where $k$ denotes the dimension of the embedding vectors, and $m$ denotes the number of distinct values in a categorical field $\{ 1, 2, \cdots, m \}$. Then, we obtain the embedding vector of $x_i^t$ by:
\begin{equation}
    \bm{v}^t_i =
\begin{cases}
\Phi_i [x_i^t]& \text{the } i\text{-th field is categorical field}\\
x_i^t \times \Phi_i& \text{the } i\text{-th field is numerical field}
\end{cases},
\end{equation}
where $\Phi_i [x_i^t]$ denotes the $x_i^t$-th row of $\Phi_i$.

\subsection{Field-level Extractor}
The field-level extractor aims to extract the event representations from the embedding vectors of fields. 
However, some existing methods~\cite{zhang2018sequential,wang2017session,tang2019towards} only use a simple dense layer and the embedding concatenation as the field-level extractor, which could not effectively extract the internal information of each behavior event. In addition, the results of the above methods are hard to interpret. We aim to design a novel field-level extractor that could not only extract the internal information of each behavior event more effective but also score the importance of different fields.

Recent studies~\cite{rendle2010factorization,wang2017deep,he2017neural,lian2018xdeepfm} find that the high-order feature interaction is very useful. Inspired by the success of factorization machines~\cite{rendle2010factorization} which captures the second-order feature interaction, we design a novel field-level extractor that captures both the first- and second-order feature interaction, and the event representations could be fed into the event-level extractor and MLP to capture higher-order feature interaction. The field-level extractor is formulated as:
\begin{equation}
    \bm{e}_t = \sum_{i=1}^n w_i^t \bm{v}_i^t + \sum_{i = 1}^n \sum_{j = i + 1}^n \bm{v}_i^t \odot \bm{v}_j^t ,\label{field-extractor}
\end{equation}
$\bm{e}_t \in \mathbb{R}^{k}$ denotes the event embedding of the $t$-th event $e_t$, and $\bm{v_i^t}$ denotes the $i$-th field embedding of the $t$-th event. $\odot$ represents Hadamard product. $w_i^t$ is the attention weight for $\bm{v_i^t}$ which is computed as:
\begin{equation}
    w_i^t = \frac{\exp{(a^i_{x_i^t})}}{ \sum_{j = 1}^n \exp{(a^j_{x_j^t})}} ,
\end{equation}
where $a^i_{x_i^t}$ is a learnable parameter for the embedding of $x_i^t$ of the $i$-th field. In other words, for each value of each categorical field and each numerical field, the model learns a learnable parameter. Note that for all values $x_i^t$ of a numerical field, the $a^i_{x_i^t}$ is the same parameter (not related to the values of $x_i^t$).
Due to all pairwise interactions need to be computed, the complexity of straight forward computation of Equation~\ref{field-extractor} is in $O(kn^2)$. Actually, it can be
reformulated to linear runtime $O(kn)$~\cite{rendle2010factorization}:
\begin{equation}
    \bm{e}_t = \sum_{i=1}^n w_i^t \bm{v}_i^t + \frac{1}{2} \left[ \left(\sum_{i=1}^n \bm{v}_i^t \right)^2 - \sum_{i=1}^n \left( \bm{v}_i^t \right) ^2 \right].
\end{equation}

Our field-level extractor is able to choose informative fields. $\bm{w}^t = [w_1^t,\cdots,w_n^t]$ indicates the attention distribution of field embeddings that could explain which field embedding is more important to represent event embedding.

\subsection{Event-level Extractor}
The event-level extractor aims to extract the sequence embedding from the historical event embedding vectors $[\bm{e}_1, \cdots, \bm{e}_{T-1}]$. Some existing approaches~\cite{he2016fusing,tang2018personalized,wang2017session,zhang2018sequential} use such as Markov chains, CNN, RNN as the event-level extractor. These methods only capture the sequential information, however, they suffer from the problem of lacking explainability. With the attention mechanism, we design an explainable event-level extractor as:
\begin{equation}
    \bm{s} = \sum_{t=1}^{T-1} u_t f_1(\bm{e}_t) ,
\end{equation}
where $u_t$ is the attention weight which is formulated as:
\begin{equation}
    \hat{u}_t = \frac{<f_2(\bm{e}_t), f_3(\bm{e}_t)>}{\sqrt{k}}, u_t = \frac{\exp{(\hat{u}_t)}}{ \sum_{j = 1}^{T-1} \exp{(\hat{u}_j)}},
\end{equation}
where $<\cdot,\cdot>$ denotes the inner product. $f_1(\cdot)$, $f_2(\cdot)$ and $f_3(\cdot)$ represent the feed-forward networks to project the input event vector $\bm{e}_t$ to one new vector representation. Actually, there are lots of ways to design $f_1(\cdot)$, $f_2(\cdot)$ and $f_3(\cdot)$. In this paper, we use one simple dense layer as $f_1(\cdot)$, $f_2(\cdot)$ and $f_3(\cdot)$.

With the attention mechanism, our event-level extractor is able to score the event importance, and from the score, we could find which event is more important to represent the sequence embedding. To analyze the score of fraud samples, we could find some high-risk behavior sequences.

\subsection{Prediction and Learning}
We concatenate the sequence embedding $\bm{s}$ and the target event embedding $\bm{e}_T$ as $[\bm{s}, \bm{e}_T]$. And then, we feed $[\bm{s}, \bm{e}_T]$ into a multi-layer perceptron (MLP) to get the final prediction $\hat{y}$:
\begin{equation}
    \hat{y} = Sigmoid\left(MLP([\bm{s}, \bm{e}_T]) + l(e_T)\right),\label{eq:lr}
\end{equation}
where $l(\cdot)$ is the linear part just like the ``wide'' part in Wide \& Deep~\cite{cheng2016wide} to capture the first-order information:
\begin{equation}
    l(e_T) = \sum_{i=1}^n c_i( x_i^T) + c_0,
\end{equation}
where $c_i(\cdot) \in \mathbb{R} (i\in \{1,\cdots,n\})$ indicates the mapping function of the $i$-th field which could denote the importance of the feature. Actually, $c_i(\cdot)$ is a look-up embedding layer as Section~\ref{lookup} with the dimension $k=1$. With the fraud sample $e_T$, it is easy to understand that the value of $c_i( x_i^T)$ should be big. Hence, the wide layer of HEN could help to identify specific value of fields with high-risk/low-risk, which could be used as whitelist/blacklist. For example, the $i$-th categorical field has two distinct values $x_1,x_2$, and $c_i(x_1) > c_i(x_2)$ indicates $x_1$ is higher-risk than $x_2$. 

For binary prediction tasks, we need to minimize the negative log-likelihood:
\begin{equation}
    L(\theta) = - \frac{1}{N} \sum_{(E, y) \in \mathcal{D} } (y \log \hat{y} + (1 - y) \log (1 - \hat{y})),\label{crossentropy}
\end{equation}
where $N$ is the number of samples, $y$ is the label of sample $e_T$, $\theta$ is the parameters set and $\mathcal{D}$ is the dataset. The overall structure of HEN is shown in Figure~\ref{fig:HEN}.

\begin{figure}[h]
	\centering
	\begin{minipage}[b]{1\linewidth}
		\centering
		\includegraphics[width=1.\linewidth]{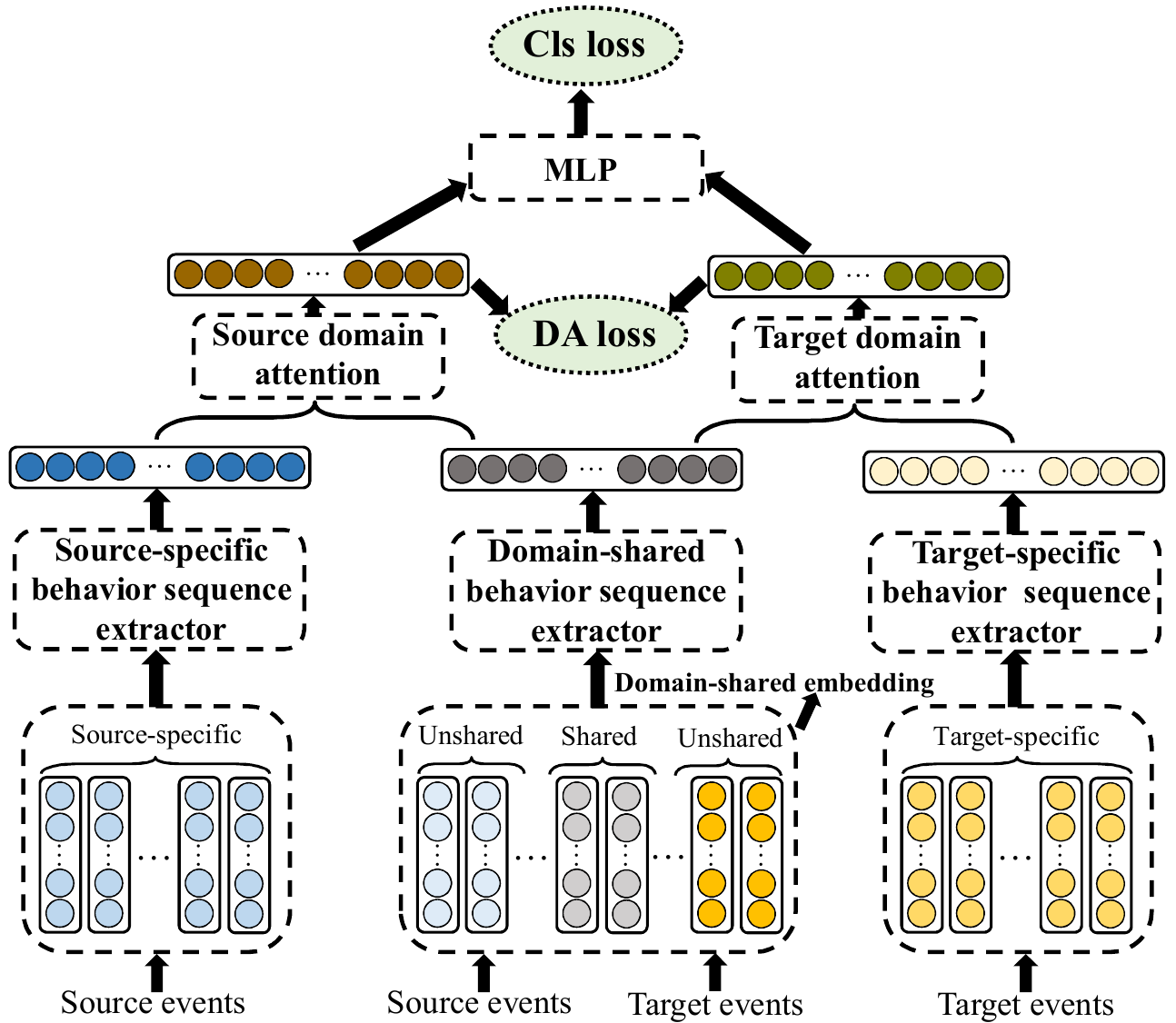}
	\end{minipage}
	\caption{The structure of the general transfer framework for cross-domain fraud detection.}\label{fig:GTF}
\end{figure}
\section{General Transfer Framework}
\subsection{Definition of Cross-domain Fraud Detection}
In cross-domain fraud detection problem, we are given a source domain $\mathcal{D}^{src} = \{(E^{src}_i, y^{src}_i)\}_{i=1}^{N^{src}}$ of $N^{src}$ labeled samples ($E^{src}_i$ has the same definition as Section~\ref{problemdef} and $y^{src}_i$ denotes the label of $e^{src}_T$ in $E^{src}_i$). Similarly, we have a target domain $\mathcal{D}^{tgt} = \{(E^{tgt}_i, y^{tgt}_i)\}_{i=1}^{N^{tgt}_{train} } \bigcup \{(E^{tgt}_i)\}_{i=1}^{N^{tgt}_{test}}$. Note that $N_{train}^{tgt} \ll N^{src}$ which is also called semi-supervised transfer problem~\cite{tzeng2015simultaneous}. The transfer task aims to improve the model performance on the unlabeled test set $\{(E^{tgt}_i)\}_{i=1}^{N^{tgt}_{test}}$ with the help of $\mathcal{D}^{src}$.

\subsection{The Strategy of Embedding}
We find that some fields of different domains have unshared values, such as IP\_address, City, while some fields share the same values, such as Event\_category, Card\_type. In addition, we conduct experiments to train a model on $\mathcal{D}^s$, and the performance of the model on $\{(E^t_i)\}_{i=1}^{N^{tgt}_{test}}$ is unsatisfying. The main reason for the unsatisfying performance is that models trained on source samples could not learn the embedding of the unshared fields.
Hence, we consider to share the embedding of the fields' shared values and learn the embedding of the unshared values respectively (named domain-shared embedding layer as shown in Figure~\ref{fig:GTF}).

With more samples, the domain-shared embedding layer could learn the shared embedding better. However, it has some negative influence. For example, even though the Issuer bank information is shared by both domains, the Issuer distribution of the fraudsters may differ a lot in the two domains. Since the source samples are much more than the target samples, the domain-shared embedding would more focus on Issuer1 which would lose the domain-specific information. Hence, we also design a source- and target-specific embedding layer to capture the domain-specific information.

\subsection{Shared and Specific behavior sequence extractor}
The behavior sequence extractor aims to extract the sequence feature from the embedding of fields for prediction. It is easy to consider sharing the behavior sequence extractor for all domain-shared, source- and target-specific embedding. However, there are two bottlenecks with single shared behavior sequence extractor: (1) The distributions of the domain-shared, source- and target-specific embedding are completely different, which may affect the performance of the network. (2) The shared behavior sequence extractor could not capture all sequential information of both domains, and it misses some domain-specific sequential information.

Hence, we propose domain-shared and domain-specific behavior sequence extractors as shown in Figure~\ref{fig:GTF}. There are three main advantages of the structure: (1) The domain-shared behavior sequence extractor focuses on common sequential information. (2) The target-specific extractor can make up for the disadvantages of the domain-shared extractor which ignores the target-specific knowledge, and it could capture sufficient target-specific sequential information. (3) The source-specific extractor could capture sufficient source-specific sequential information. Here, we formulate the representations extracted by domain-shared, target-specific and source-specific extractors as $\bm{z}_{share}, \bm{z}^{tgt}_{spe}, \bm{z}^{src}_{spe}$, respectively. 

\subsection{Domain Attention}
The shared behavior sequence extractor could extract sequence representations that contain domain-shared sequential information. In addition, the domain-specific behavior sequence extractor could extract more specific sequential information which could make up for the disadvantages of the domain-shared extractor. By combining the two representations $\bm{z}_{share}, \bm{z}_{spe}$, it could contain more useful knowledge. The simplest methods to combine the two representations are such as average and concatenation. However, for different samples, the weights of the shared and specific knowledge would be different. Meanwhile, they could not explain from which part the knowledge is more important for target prediction. Hence, we also propose a domain attention mechanism to combine the two representations:
\begin{equation}
    \bm{z}^o = b^o_{share} g_1(\bm{z}^o_{share}) + b^o_{spe} g_1(\bm{z}^o_{spe}),
\end{equation}
where $o \in \{tgt, src\}$ denotes the domain of the representations $\bm{z}$, and $b^o$ is the attention weight which is formulated as:
\begin{equation}
\begin{split}
    \hat{b}^o_{p} = \frac{<g_2(\bm{z}^o_{p}), g_3(\bm{z}^o_{p})>}{\sqrt{k}}, 
    b^o_{p} = \frac{\exp{(\hat{b}^o_{p})}}{ \exp{(\hat{b}^o_{share})} + \exp{(\hat{b}^o_{spe})}},
\end{split}
\end{equation}
where $p \in \{share, spe\}$, $g_1(\cdot)$, $g_2(\cdot)$ and $g_3(\cdot)$ are the feed-forward networks to project the input representations $\bm{z}$ to new representations. Domain attention module is capable of learning the importance of domain-shared and domain-specific representations.

\subsection{Aligning Distributions}
Note that the distributions of representations $\bm{z}^{src}$ and $\bm{z}^{tgt}$ are different. We could feed them into a source MLP and a target MLP, respectively. However, we find the performance is unsatisfying, and the reason would be the limitation of target samples which lead to the target MLP overfitting. Thus we consider aligning the distributions of representations $\bm{z}^{src}$ and $\bm{z}^{tgt}$, and feeding them into the same MLP to avoid overfitting. Most existing transfer approaches~\cite{long2015learning,tzeng2015simultaneous,sun2016deep,ganin2016domain,xie2018learning} aim to align the marginal and conditional distributions. However, in our scenario, the class distribution is extremely unbalanced that the number of non-fraud samples is about 100 times more than fraud samples. Hence, aligning the marginal and conditional distributions would lead to unsatisfying performance. For our scene, we propose Class-aware Euclidean Distance which explicitly takes the class information into account and measures the intra-class and inter-class discrepancy across domains. First, we formulate Euclidean Distance between two classes as:
\begin{equation}
    d(\mathcal{D}^{o_1}_{c_1}, \mathcal{D}^{o_2}_{c_2}) = \left \|  \frac{1}{N^{o_1}_{c_1}}\sum_{i = 1}^{N^{o_1}_{c_1}} \bm{z}^{o_1}_i - \frac{1}{N^{o_2}_{c_2}}\sum_{i = 1}^{N^{o_2}_{c_2}} \bm{z}^{o_2}_i  \right \|^2,
\end{equation}
where $c_1$ and $c_2$ denote the class label (fraud and non-fraud), and $N^{o_1}_{c_1}, N^{o_2}_{c_2}$ represent the numbers of samples of classes $c_1, c_2$ in $o_1$ and $o_2$ domains, respectively. The $d(\mathcal{D}^{o_1}_{c_1}, \mathcal{D}^{o_2}_{c_2})$ denotes the Euclidean Distance between the embeddings of class $c_1$ in $o1$ domain and class $c_2$ in $o_2$ domain. Note that $o_1$ and $o_2$ could denote the same domain. Then, we formulate Class-aware Euclidean Distance as:
\begin{equation}
    \Delta(\mathcal{D}^{src}, \mathcal{D}^{tgt}) = \frac{\sum_{c \in \{0,1\}} d(\mathcal{D}^{src}_{c}, \mathcal{D}^{tgt}_{c})}{ \sum_{o_1 \in \{src, tgt\}} \sum_{o_2 \in \{src, tgt\}} d(\mathcal{D}^{o_1}_{c=0}, \mathcal{D}^{o_2}_{c=1}) },\label{CED}
\end{equation}
where the numerator represents the sum of intra-class distance and the denominator denotes the sum of inter-class distance. Existing transfer methods that align marginal distributions ignore the class information~\cite{long2015learning,tzeng2015simultaneous,sun2016deep}. The transfer methods which match conditional distributions consider the intra-class information without considering the inter-class information~\cite{xie2018learning}. By minimizing the left part of Equation (\ref{CED}), the intra-class domain discrepancy is minimized to compact the feature representations of samples within a class, whereas the inter-class domain discrepancy is maximized to push the representations of each other further away from the decision boundary. Hence, Class-aware Euclidean Distance could achieve better performance in our scene.

\subsection{Apply The General Transfer Framework upon Various Models}
The general transfer framework as shown in Figure~\ref{fig:GTF} could be applied upon various existing models in the Embedding \& MLP paradigm. The most important thing to apply the transfer framework is to define the behavior sequence extractor. For our HEN, the behavior sequence extractor contains a field-level extractor and an event-level extractor as shown in Figure~\ref{fig:HEN}. For neural factorization machines (NFM)~\cite{he2017neural}, we use the FM module as the behavior sequence extractor. For wide \& deep~\cite{cheng2016wide}, we use a dense layer as the behavior sequence extractor. For all models, we feed the output of the behavior sequence extractor into an MLP to get the final prediction. Other details are easy to understand as shown in Figure~\ref{fig:GTF}.

The training mainly follows the back-propagation algorithm. The target loss is formulated as:
\begin{equation}
    L = L_{cls} + \lambda L_{da},
\end{equation}
where $\lambda$ is a trade-off parameter, $L_{cls}$ denotes the classification loss and $L_{da}$ denotes the domain adaptation loss. In this paper, we choose the Class-aware Euclidean Distance in Equation (\ref{CED}) as $L_{da}$. $L_{cls}$ can choose most classification loss, such as negative log-likelihood, Least Square loss, Pair-wise loss. In this paper, we adopt negative log-likelihood for training as Equation (\ref{crossentropy}).

\section{Experiment}\label{experiment}
\subsection{Datasets}

\begin{table}[!t]
  \centering
  \caption{The statistics of datasets. \#pos and \#neg mean the number of fraud and non-fraud samples. \#pos ratio = $\frac{\mathtt{\#pos}}{\mathtt{\#pos} + \mathtt{\#neg}}$. \#fields and \#events mean the number of fields and events, respectively. \#avglen represents the average length of historical event sequences.}
   \setlength{\tabcolsep}{1mm}{
    \begin{tabular}{ccccccc}
    \toprule
    Dataset & \#pos & \#neg & \#pos ratio & \#fields &  \#events & \#avglen \\
    \midrule
     C1    & 10K   & 1.93M & 5.2\textperthousand & 56 & 3.57M & 4.94 \\
     C2    & 4.2K  & 1.48M & 2.8\textperthousand & 56 & 3.91M & 4.73 \\
     C3    & 15K   & 1.37M & 10.8\textperthousand & 56 & 4.28M & 14.97 \\
     C4    & 5.7K  & 174K  & 31.7\textperthousand & 56 & 353K & 4.91 \\
    \bottomrule
    \end{tabular}%
    }
  \label{tab:dataset}%
\end{table}%

\begin{figure}[t!]
	\centering
	\begin{minipage}[b]{1\linewidth}
		\centering
		\includegraphics[width=.95\linewidth]{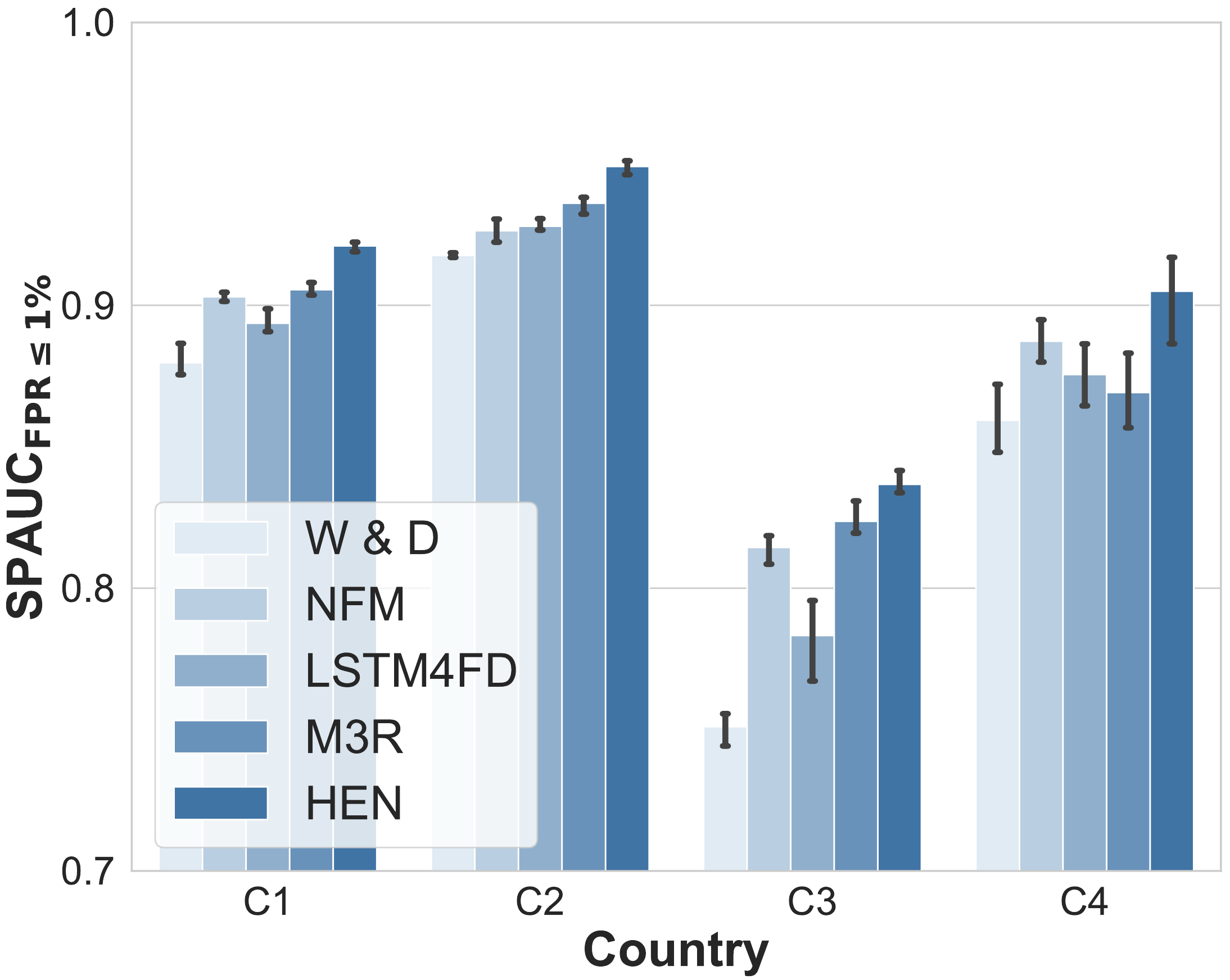}
	\end{minipage}
	\caption{Supervised classification $\mathbf{SPAUC_{FPR \leq 1\%}}$ on four countries.}\label{fig:supervisedall}
\end{figure}

The fraud detection dataset~\footnote{In order to comply with the data protection regulation in each country, multiple approaches have been taken in the data processing step, which include but are not limited to: personally identifiable information (PII) encrypted with salted MD5, data abstraction, data sampling, etc. By doing so, no original data could be restored and the statistics in this manuscript does not represent any the real business status. Meanwhile the dataset is generated only for research purpose in our study and will be destroyed after the experiments.} is collected from one of the world-leading cross-border e-commerce company, which utilizes the risk management system to detect the transaction frauds.
The dataset contains the card transaction samples from four countries (\textbf{C1},\textbf{ C2}, \textbf{C3}, \textbf{C4}),  across a timespan of 10 weeks in 2019.
For samples in each country, we utilize users' historical behavior sequences of the last month. The task is to detect whether the current payment event is a card-stolen case, with knowing users' historical sequential information. The fraud labels are collected from the chargeback reports from card issuer banks (e.g., the card issuer receives claims on unauthorized charges from the cardholders and report related transaction frauds to the merchants) and label propagation (e.g., the device and card information are also utilized to mark similar transactions). The details of the dataset are listed in Table~\ref{tab:dataset}. 
%Then, we sort all users' prediction events according to timestamp order, taking the first $80\%$ as training set, the following $10\%$ for the validation set and the remaining $10\%$ for the test set. These are consistent for all models for fair comparison.

\subsection{Base Models}
To show the effectiveness of HEN, we choose 4 prediction models as baselines.

\begin{itemize}
\item \textbf{W \& D~\cite{cheng2016wide}:} In real industrial applications, Wide \& Deep model has been widely accepted. It consists of two parts: i) wide model, which handles the manually designed cross-field features, ii) deep model, which automatically extracts nonlinear relations among features. 

\item \textbf{NFM~\cite{he2017neural}:} It is a recent state-of-the-art simple and efficient neural factorization machine model. It feeds dense embedding into FM, and the output of FM is fed to MLP for capturing higher-order feature interactions.

\item \textbf{LSTM4FD~\cite{zhang2018sequential,wang2017session}:} Some work has applied LSTM for the sequence-based fraud detection tasks, and we called these methods as LSTM4FD. %For LSTM4FD, we compute the average of fields' embeddings as event representations and then apply the event representations into a LSTM. Finally, the output of LSTM is fed into MLP to get the prediction.

\item \textbf{M3R~\cite{tang2019towards}:} It is a most recent hierarchical sequence-based model (M3R and M3C) which deals with both short-term and long-term dependencies with mixture models. We choose the better hierarchical model M3R as the baseline.

\end{itemize}

Note that our transfer framework is a general framework that can be applied upon various existing models in the Embedding \& MLP paradigm. To show the compatibility of our transfer framework, we apply our transfer framework upon both non-hierarchical models (W \& D, NFM) and hierarchical models (LSTM4FD, M3R, HEN). For W \& D, we add a dense layer as the behavior sequence extractor before MLP. For NFM, we use the FM as the extractor before MLP. For hierarchical models, we use the combination of event-level and field-level extractor as the behavior sequence extractor.

\begin{figure}[t!]
	\centering
	\begin{minipage}[b]{1\linewidth}
		\centering
		\includegraphics[width=.95\linewidth]{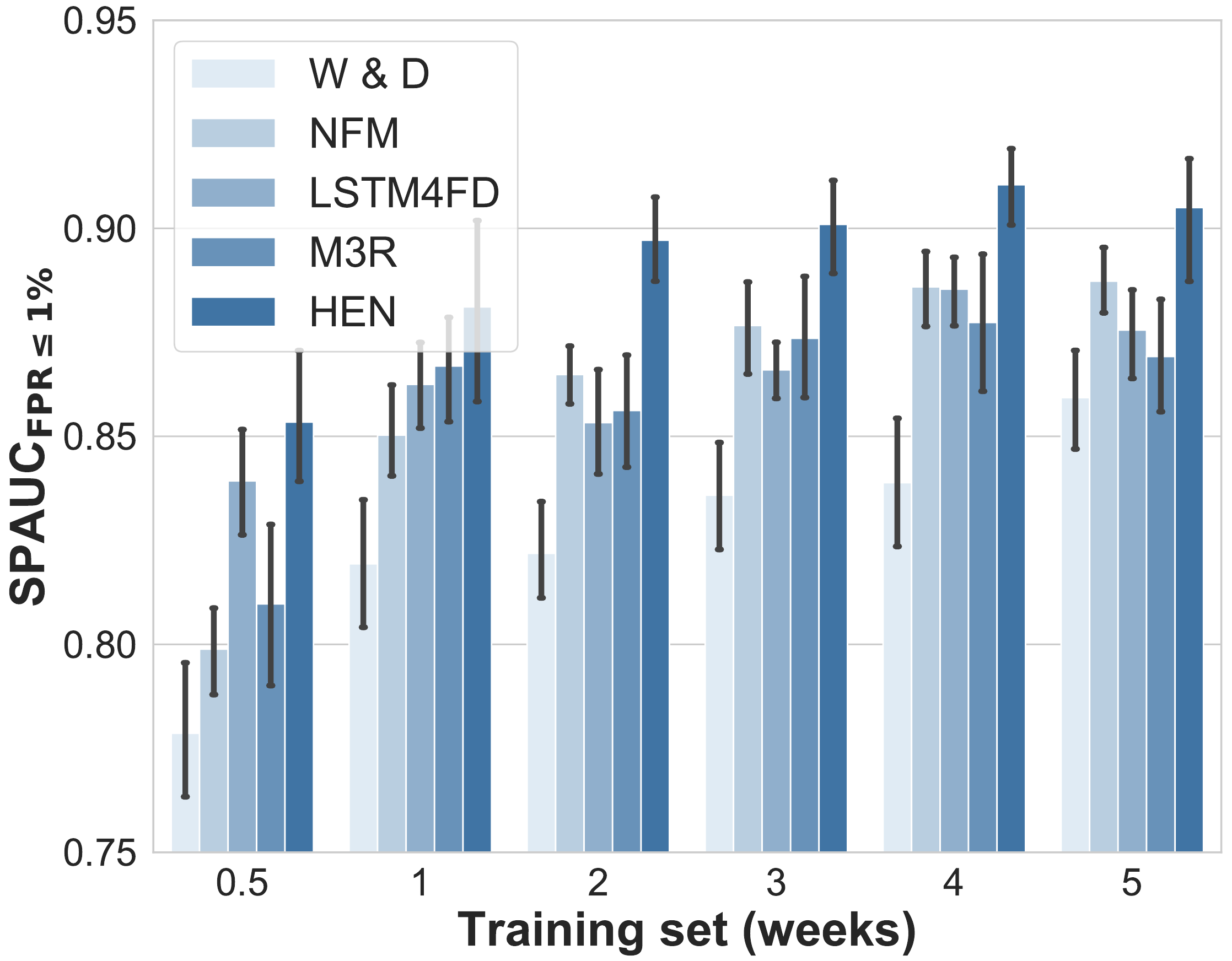}
	\end{minipage}
	\caption{Supervised classification $\mathbf{SPAUC_{FPR \leq 1\%}}$ on C4 with different split of the dataset C4.}\label{fig:supervisedVN}
\end{figure}

\subsection{Experimental Set-Up}
\subsubsection{Dataset splits}
The dataset contains the card transaction samples from four countries across the same time span of 10 weeks in 2019. Then, for each country, we sort all users' prediction events according to timestamp order, taking the first 5 weeks as the training set, the following 2 weeks as the validation set and the remaining 3 weeks as the test set. To demonstrate the effectiveness of HEN, we run standard supervised prediction experiments on the dataset of four countries.

The e-commerce of the company expands to C4 recently, and it is easy to find that the size of the C4 dataset is much smaller than the other three countries. Hence, we use C4 as the target domain and the other three countries (C1, C2, C3) as source domains. To prove our transfer framework is widely applicable, we use [0.5, 1, 2, 3, 4, 5] weeks of the training set of each country as new training sets (six divisions) with the validation and test sets unchanged. For transfer experiments, we use the same divisions of source and target training sets to train the model and then evaluate the performance on the target domain.

\subsubsection{Evaluation metric.}
In binary prediction tasks, AUC (Area Under ROC) is a widely used metric~\cite{fawcett2006introduction}. However, in our real card-stolen fraud detection
scenario, we should increase the recall rate, while avoid disturbing the normal users as few as possible. In other words, the task is improving the True Positive Rate (TPR) on the basis of low False Positive Rate (FPR). In our scenario, we should pay attention to partial AUC (\textbf{AUC}$_\mathbf{FPR \leq maxfpr}$) which denotes the area of the head of the ROC curve when the $\mathrm{FPR}\leq \mathrm{maxfpr}$. While the $\mathrm{maxfpr}$ is very low, AUC$_{\mathrm{FPR} \leq \mathrm{maxfpr}}$ has a small range of variation which makes it difficult to compare the performance of the model. Therefore, we adopt the standardized partial AUC (\textbf{SPAUC}$_\mathbf{FPR \leq maxfpr}$)~\cite{mcclish1989analyzing}:
\begin{equation}
\begin{split}
    \mathrm{SPAUC_{FPR \leq maxfpr}} &= \frac{1}{2} \left (1 + \frac{ \mathrm{AUC_{FPR \leq maxfpr} } - \mathrm{minarea}}{\mathrm{maxarea} - \mathrm{minarea}}\right ), \\
    \mathrm{where} \quad \mathrm{maxarea} &= \mathrm{maxfpr},\\
    \mathrm{minarea} &= \frac{1}{2} \times \mathrm{maxfpr}^2.
\end{split}
\end{equation}
It is easy to understand the range of \textbf{SPAUC}$_\mathbf{FPR \leq maxfpr}$ is 0.5 to 1 (we assume the prediction by models is better than stochastic prediction). In practice, we require FPR to be less than 1\%. Hence, in this paper, we use $\mathrm{SPAUC_{FPR \leq 1\%}}$ for all experiments.

\begin{figure*}[t!]
	\centering
	\begin{minipage}[b]{1\linewidth}
		\centering
		\includegraphics[width=.95\linewidth]{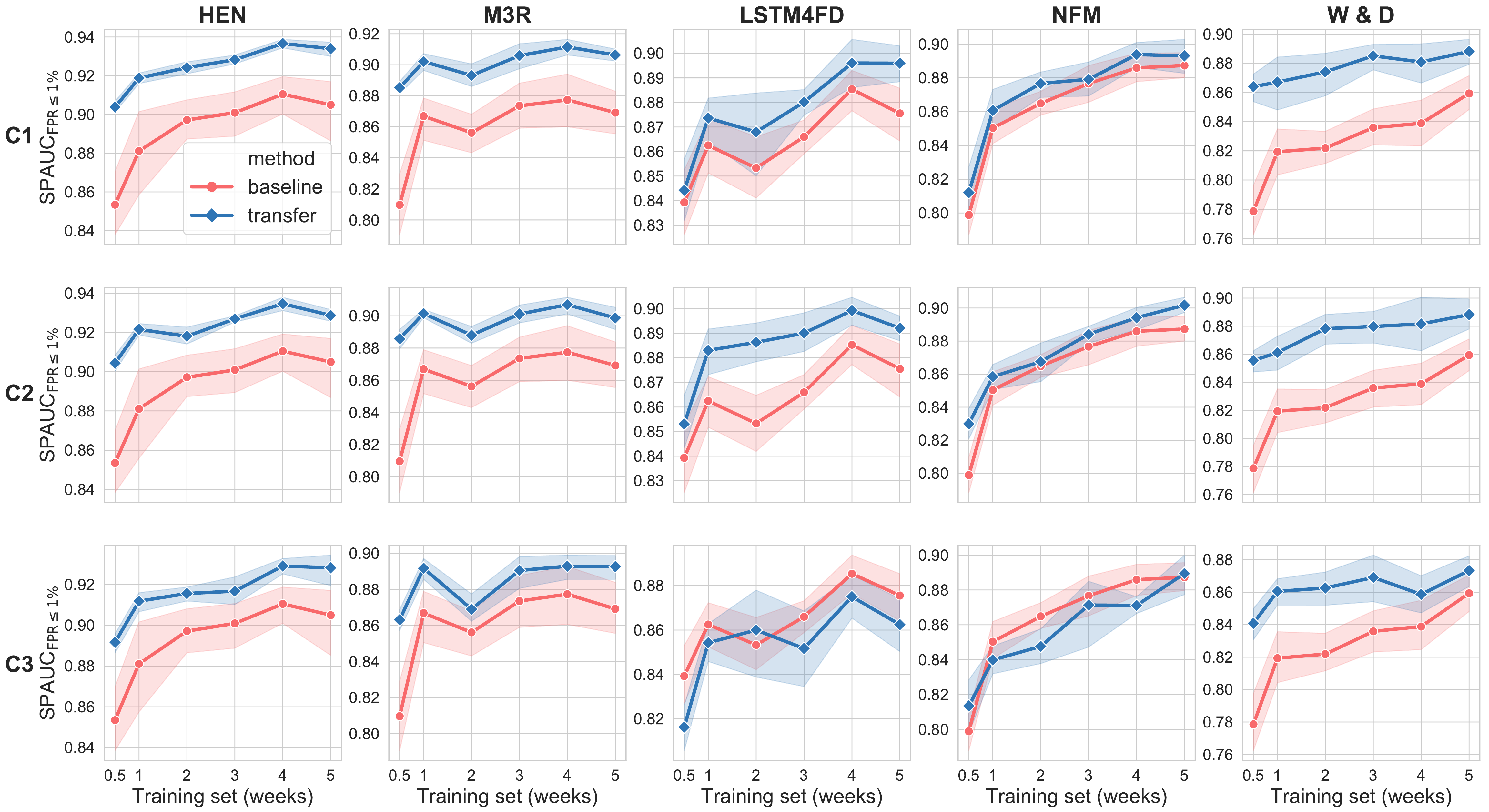}
	\end{minipage}
	\caption{The results ($\mathbf{SPAUC_{FPR \leq 1\%}}$) of the transfer experiments on 3 (source countries) $\times$ 5 (models) $\times$ 6 (divisions) $=$ 90 tasks.}\label{fig:transfer_experiments}
\end{figure*}

\subsubsection{Implementation  Details}
For a fair comparison, we use the same setting for all methods. The MLP in these models use the same structure with two dense layers (hidden units 64), and the LSTM in both LSTM4FD and M3R is the Bi-LSTM~\cite{graves2013speech} with a single layer. The dimensionality of embedding vectors of each input field is fixed to 16 for all our experiments. In addition, we set: learning rate of 0.005, maximum number of events $T = 10$, dropout (keep probability 0.8). Besides, there is a clear class imbalance in the dataset as shown in Table~\ref{tab:dataset}, so we upsample the positive sample to 5 times. For non-hierarchical models, we combine all the features of user's events (history events and the current payment event) as the input. Following~\cite{ganin2016domain,long2015learning}, instead of fixing the adaptation factor $\lambda$, we gradually change it from $0$ to $1$ by a progressive schedule: $\lambda_\theta = \frac{2}{exp(-10 \theta)} - 1$, and $\theta$ denotes the training step. We do not perform any datasets-specific tuning except early stopping on validation sets. Training is done through stochastic gradient descent over shuffled mini-batches with the Adam~\cite{kingma2014adam} update rule. For each task, we report the \textbf{SPAUC}$_\mathbf{FPR \leq maxfpr}$ and 95\% confidence intervals on ten random trials.

\subsection{Results}
\subsubsection{Standard supervised prediction tasks} 

We demonstrate the effectiveness of our HEN on standard supervised prediction tasks of two parts. (1) We run experiments on the four countries with 5 weeks of data as training sets, and the results are shown in Figure~\ref{fig:supervisedall}. (2) We conduct experiments on the C4 with training sets of six different divisions, and the results are shown in Figure~\ref{fig:supervisedVN} (the results are the baseline of transfer tasks). The experimental results further reveal several insightful observations.
\begin{itemize}
    \item With different countries or different divisions, HEN outperforms all compared methods which demonstrates the effectiveness of HEN. The improvement mainly comes from the hierarchical structure and higher-order feature interactions.
    \item Hierarchical models could achieve better performance on sequence-based fraud detection tasks. For the models with higher-order feature interactions, HEN which is a hierarchical model outperforms the non-hierarchical models NFM. For the models only considering the first-order feature, hierarchical LSTM4FD and M3R outperform non-hierarchical W \& D on most tasks.
    \item Considering higher-order feature interactions could improve the performance. Comparing non-hierarchical models W \& D and NFM, NFM which considers higher-order feature interactions achieves a better result. The comparison results among HEN and LSTM4FD, M3R could draw the same conclusion.
    \item On the tasks of C1, C2, C3, M3R achieves better results except for HEN. However, on most tasks of C4, the performance of M3R is worse than NFM, LSTM4FD. We conjecture that the field-level of M3R is a dense layer that would be overfitting on small datasets. 
\end{itemize}
%Overall, all the above results demonstrate the effectiveness of the HEN.

\subsubsection{Transfer tasks}
Our transfer framework is a general framework that can be applied upon various existing models in the Embedding \& MLP paradigm. Thus we apply the transfer framework on HEN, M3R, LSTM4FD, NFM, W \& D. Besides, we use three counties (C1, C2, C3) as the source domains while the country C4 which has much fewer samples as the target domain. In order to prove that our transfer framework is applicable to different sizes of training sets, we use the six divisions of training sets (\{0.5, 1, 2, 3, 4, 5\} weeks). Note that the training sets of source and target domains should be in the same time period, e.g., both C1 and C4 use 3 weeks training samples as the training set. We conduct transfer experiments on 90 tasks ($3 \times 5 \times 6 = 90$, 3 source countries, 5 base models, 6 divisions). The results of the transfer experiments are shown in Figure~\ref{fig:transfer_experiments}, and the red lines are the baselines which only use the target training set to train model, also shown in Figure~\ref{fig:supervisedVN}. From the results, we have the following findings:

\begin{table*}[htbp]
  \centering
  \caption{Results ($\mathbf{SPAUC_{FPR \leq 1\%}}$) of ablation study on transfer tasks from dataset C1 to dataset C4 based with HEN.}
    \begin{tabular}{cccccccc}
    \toprule
    \textbf{Methods} & \textbf{3days} & \textbf{1week} & \textbf{2weeks} & \textbf{3weeks} & \textbf{4weeks} & \textbf{5weeks} & \textbf{avg} \\
    \midrule
    \textbf{target} & 0.8534$\pm$0.0173 & 0.8811$\pm$0.0232 & 0.8971$\pm$0.0111 & 0.9009$\pm$0.0117 & 0.9105$\pm$0.0099 & 0.9050$\pm$0.0171 & 0.8913 \\
    \textbf{source} & 0.5724$\pm$0.0230 & 0.5732$\pm$0.0272 & 0.5909$\pm$0.0576 & 0.5877$\pm$0.0334 & 0.5474$\pm$0.0393 & 0.5705$\pm$0.0419 & 0.5737 \\
    \midrule
    \textbf{pretrain} & 0.8639$\pm$0.0047 & 0.8920$\pm$0.0066 & 0.8983$\pm$0.0115 & 0.9120$\pm$0.0051 & 0.9172$\pm$0.0064 & 0.9191$\pm$0.0054 & 0.9004 \\
    \textbf{domain-shared} & 0.8811$\pm$0.0110 & 0.8995$\pm$0.0119 & 0.9025$\pm$0.0153 & 0.9157$\pm$0.0109 & 0.9204$\pm$0.0068 & 0.9194$\pm$0.0067 & 0.9065 \\
    \textbf{our structure} & 0.8801$\pm$0.0101 & 0.9071$\pm$0.0042 & 0.9038$\pm$0.0085 & 0.9152$\pm$0.0094 & 0.9226$\pm$0.0081 & 0.9233$\pm$0.0113 & 0.9087 \\
    \midrule
    \textbf{Coral} & 0.8566$\pm$0.0156 & 0.8938$\pm$0.0096 & 0.9028$\pm$0.0112 & 0.9163$\pm$0.0128 & 0.9236$\pm$0.0080 & 0.9221$\pm$0.0081 & 0.9025 \\
    \textbf{Adversarial} & 0.8763$\pm$0.0113 & 0.8963$\pm$0.0135 & 0.8951$\pm$0.0109 & 0.9116$\pm$0.0099 & 0.9289$\pm$0.0053 & 0.9260$\pm$0.0069 & 0.9057 \\
    \textbf{MMD} & 0.8744$\pm$0.0107 & 0.8964$\pm$0.0105 & 0.9049$\pm$0.0091 & 0.9185$\pm$0.0058 & 0.9249$\pm$0.0075 & 0.9213$\pm$0.0057 & 0.9067 \\
    \textbf{CMMD} & 0.8797$\pm$0.0136 & 0.8897$\pm$0.0132 & 0.8963$\pm$0.0145 & 0.9121$\pm$0.0135 & 0.9219$\pm$0.0069 & 0.9146$\pm$0.0122 & 0.9024 \\
    \textbf{ED} & 0.8767$\pm$0.0148 & 0.9010$\pm$0.0088 & 0.9180$\pm$0.0063 & 0.9186$\pm$0.0073 & 0.9258$\pm$0.0076 & 0.9251$\pm$0.0046 & 0.9109 \\
    \textbf{CED} & 0.8866$\pm$0.0091 & 0.9076$\pm$0.0073 & 0.9140$\pm$0.0059 & 0.9220$\pm$0.0070 & 0.9305$\pm$0.0076 & 0.9298$\pm$0.0059 & 0.9151 \\
    \textbf{ours} & \textbf{0.9038}$\pm$0.0052 & \textbf{0.9188}$\pm$0.0052 & \textbf{0.9242}$\pm$0.0063 & \textbf{0.9283}$\pm$0.0039 & \textbf{0.9367}$\pm$0.0041 & \textbf{0.9340}$\pm$0.0059 & \textbf{0.9243} \\
    \bottomrule
    \end{tabular}%
  \label{tab:ablation}%
\end{table*}%

\begin{itemize}
    \item On most tasks, our transfer framework is effective to improve the performance of the base models which also demonstrates the transfer framework can be applied upon various existing models in the Embedding \& MLP paradigm. 
    \item  Comparing with C1 and C2 as source domains, the performance of the transfer framework with C3 as the source domain is unsatisfying. The reason would be that C1 and C2 are more similar to the target country C4, while C3 is more different from C4. As shown in Table~\ref{tab:dataset}, the average length of the historical behavior sequences of C3 is 14.97, while the other three countries are about 4.8.
    \item On the tasks with C3 as source domain, the transfer framework is effective for HEN, M3R, W \& D, while it is unsatisfying for LSTM4FD, NFM. We think that the behavior extractors of HEN, M3R, W \& D contain dense layers that have a stronger fitting ability to fit the dissimilar source domain.
    \item The improvement on NFM is smaller than the other four base models. The main reason would be that the user behavior extractor of NFM is a non-parametric FM that is hard to fit the various distributions of different domains. On the contrary, the user behavior extractors of HEN, M3R, LSTM4FD, W \& D are more complex parametric layers that are easy to adapt to the different distributions.
\end{itemize}
Overall, we observe that the transfer framework is able to improve the performance of base models with different sizes of training sets, which proves that the transfer framework is compatible with many models in the Embedding \& MLP paradigm.

\subsection{Ablation Study}
To demonstrate how each component contributes to the overall performance, we now present an ablation test on our transfer framework. 
To prove the effectiveness of the transfer framework, we not only compare each component but also compare it with some baselines. We divide the ablation study into three parts: (1) Only use samples of single domain: target (only use data of target domain), source (only use data of source domain). (2) Considering the design of structure: pretrain~\cite{hinton2006fast} (first train a model on source domain and then fine-tune on target domain), domain-shared (combine source and target dataset into single dataset, and share the common embedding and network), our structure (the structure designed by us contains domain-shared, domain-specific parts and the domain attention without domain adaptation loss). (3) Based on our structure, compare different domain adaptation loss: Coral~\cite{sun2016deep} (align the second-order statistics of the source and target distributions), Adversarial~\cite{ganin2016domain} (adversarial training), MMD~\cite{long2015learning,zhu2019aligning} (a kernel two-sample test), CMMD~\cite{long2013transfer,wang2017balanced,zhu2019multi} (conditional MMD), ED (Euclidean Distance), CED~\cite{xie2018learning} (Conditional Euclidean Distance), ours (use the Class-aware Euclidean Distance to align the distributions, which also denotes the overall transfer framework).

\begin{table}[h]
  \centering
  \caption{The extracted high-risk (high weight) and low-risk (low weight) features according to the learned weights of the wide layer in Equation~\ref{eq:lr}, the ``Email\_suffix'' is suffix of the user-bound Email, the ``Card\_bin''  is the last six digits of the card number, the ``Issuer'' is issuer name.}
    \begin{tabular}{clll}
    \toprule
          & Email\_suffix & Card\_bin & Issuer \\
    \midrule
   \multicolumn{1}{c}{\multirow{3}[0]{*}{\shortstack{High-\\Risk}}}  & email1(123/131) & card1(33/33) & issuer1(602/607) \\
         & email2(27/27) & card2(42/54) & issuer2(76/108) \\
          & email3(54/59) & card3(77/78) & issuer3(77/90) \\
    \midrule
    \multicolumn{1}{c}{\multirow{3}[2]{*}{\shortstack{Low-\\Risk}}} & email4(0/382) & card4(0/2365) & issuer4(27/12491) \\
          & email5(0/298) & card5(0/245) & issuer5(0/789) \\
          & email6(0/471) & card6(0/5972) & issuer6(1/725) \\
    \bottomrule
    \end{tabular}%
  \label{tab:caseWL}%
\end{table}%

\begin{figure*}[t!]
	\centering
	\centering
	\begin{minipage}[b]{1\linewidth}
		\centering
		\includegraphics[width=.9\linewidth]{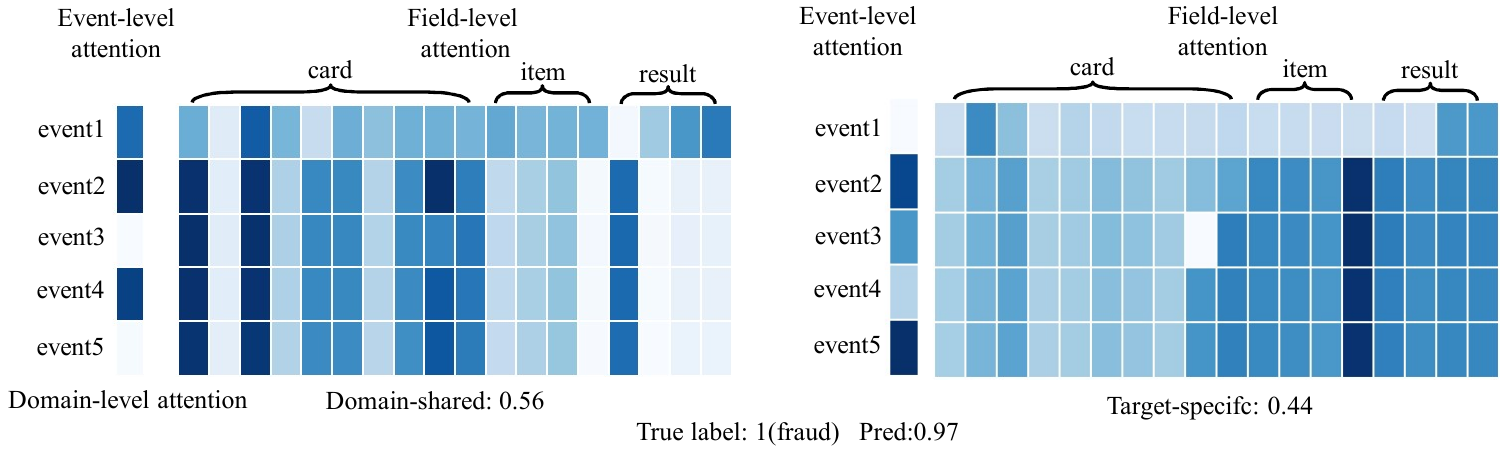}
	\end{minipage}
	\caption{Case study: we present three-level attention. Note that we only analyze the abnormal point of history sequence. Each column of field-level attention means a field, e.g., the card-related fields contain such as card\_bin, card expiration time, issuer.}\label{fig:case}
\end{figure*}

We conduct experiments for ablation study with dataset C1 as the source domain and dataset C4 as the target domain. Besides, all experiments are based on HEN, and the results are shown in Table~\ref{tab:ablation}. From the results, we can make interesting observations:
\begin{itemize}
    \item The performance of `our structure' is better than `target', which proves the domain-shared and domain-specific structure is useful. `ours' outperforms `our structure' that proves the aligning distributions with Class-aware Euclidean Distance is effective. Hence, each component of the transfer framework is useful and effective.
    \item In the first part, the performance of `source' is largely worse than `target' which proves the domain shift between source and target domains could seriously destroy the performance. Thus it is necessary to use effective transfer methods.
    \item Comparing `our structure' with `pretrain' and `domain-shared', we could find `our structure' is more effective. `pretrain' and `domain-shared' are simple and generally adopted transfer methods. The results show `ours' outperforms the general transfer methods which demonstrate our transfer framework is more suitable for cross-domain fraud detection.
    \item Comparing different domain adaptation loss, we could find `ours' outperforms both marginal and conditional methods which prove the effectiveness of Class-aware Euclidean Distance in this scene. Besides, the performance of `Coral', `Adversarial', `MMD', `CMMD' is worse than `our structure' without domain adaptation loss, which reveals that the inappropriate aligning methods could lead to negative transfer.
\end{itemize}

Overall, we observe each component of the transfer framework is effective, and the Class-aware Euclidean Distance is more suitable than previous marginal and conditional aligning methods for our scenario.

\subsection{Case Study}
\subsubsection{Explanation of Wide Layer}
Firstly, we extract some high-risk and low-risk features according to the learned weights of the wide layer in Equation (\ref{eq:lr}). The corresponding features are listed in Table~\ref{tab:caseWL}. Since the email\_suffix, card\_bin and issuer name need to be kept private, we only use such as email1, card1, issuer1 to denote the ID and name. For each field, we select the three highest-risk and lowest-risk samples with specific values and present the ratio of each feature (black number / total number) which is helpful to understand the reason for risks. From Table~\ref{tab:caseWL}, we find the wide layer could learn the blacklist and whitelist.

\subsubsection{Three-level Attention} HEN with the general transfer framework has three-level attention and each level has its own explanation settings. Field-level attention could help us understand which field is important for the target prediction. Event-level attention could find the important events, and from the important events, we could find the high-risk sequences. The domain-level attention could show the importance of domain-shared and domain-specific knowledge. We present one case with explanations of positive (fraud) samples in Figure~\ref{fig:case} on task (C1 $\rightarrow$ C4). To focus on showing the capability of the three-level explanation, we only present the case about the historical behavior sequences. In addition, we use 56 fields in our experiments, but only show the representative fields in Figure~\ref{fig:case}.

The case in Figure~\ref{fig:case} is a typical fraud case and the prediction of the sample is 0.97. The five history events are 1 register event and 4 payment events. (left) The domain-shared attention is 0.56. Besides, the domain-shared extractor more focuses on event1,2,4 and card-related fields. Then we analyze the history events: after registration, the fraudster immediately used a high-risk card\_bin which had been used to fraud 5 times. For the event4, the fraudster tried to change the card expiration time to make the system recognize it as a new card. (right) The target-specific attention is 0.44. The target-specific extractor more focuses on event2-5 and item- and result-related fields. Then we analyze the history events: the fraudster tried to buy the same item 4 times. All of the four payment events triggered 3D verification and failed. From the case, we could find domain-shared and target-specific extractors would focus on different fields and different behavior sequences.

\section{Conclusion}
\label{sec:conclusion}

In this paper, we studied the online transaction fraud detection problem from the perspective of modeling users' behavior sequences. Along this line, to effectively capture the sequential information and explore the explainability for fraud detection, we proposed a Hierarchical Explainable Network (HEN). HEN can extract both representations of the target behavior events and users' historical behavior sequences to significantly improve prediction performance. Furthermore, to handle the real-world scenarios when there are only limited labeled data in the target domain with hardly model reused, we proposed a general transfer framework that can not only be applied upon HEN but also various existing models in the Embedding \& MLP paradigm. Finally, we conducted extensive experiments on real-world data sets collected from a world-leading cross-border e-commerce platform to validate the effectiveness of our proposed models, and the case study further approves the explainability of our models.

\begin{acks}
The research work is supported by the National Key Research and Development Program of China under Grant No. 2018YFB1004300, the National Natural Science Foundation of China under Grant No. U1836206, U1811461, 61773361, the Project of Youth Innovation Promotion Association CAS under Grant No. 2017146. This work is funded in part by Ant Financial through the Ant Financial Science Funds for Security Research. We also thank Minhui Wang, Zhiyao Chen, Changjiang Zhang, Dan Hong for their valuable suggestions.
\end{acks}

\balance
\bibliographystyle{ACM-Reference-Format}
\bibliography{main}

\end{document}